\documentclass[10pt,conference]{IEEEtran}
\usepackage{graphicx}
\usepackage{listings}
\usepackage{hyperref}
\usepackage{subfig}
\usepackage{cite}
\usepackage{float}
\usepackage{booktabs}
\usepackage{tikz}
\usetikzlibrary{arrows}
\usetikzlibrary{calc}
\usepackage{enumitem}

\title{LLM Performance for Code Generation \\on Noisy Tasks}

\author{
    Radzim Sendyka, Christian Cabrera, Andrei Paleyes, Diana Robinson, Neil Lawrence \\
    {University of Cambridge} \\
    \texttt{\{rs2071, chc79, ap2169, dmpr3, ndl21\}@cam.ac.uk}
}

\date{May 2025}

\begin{document}

\maketitle

\begin{abstract}
This paper investigates the ability of large language models (LLMs) to recognise and solve tasks which have been obfuscated beyond recognition. Focusing on competitive programming and benchmark tasks (LeetCode and MATH), we compare performance across multiple models and obfuscation methods, such as noise and redaction. We demonstrate that all evaluated LLMs can solve tasks obfuscated to a level where the text would be unintelligible to human readers, and does not contain key pieces of instruction or context. We introduce the concept of eager pattern matching to describe this behaviour, which is not observed in tasks published after the models’ knowledge cutoff date, indicating strong memorisation or overfitting to training data, rather than legitimate reasoning about the presented problem. We report empirical evidence of distinct performance decay patterns between contaminated and unseen datasets. We discuss the implications for benchmarking and evaluations of model behaviour, arguing for caution when designing experiments using standard datasets. We also propose measuring the decay of performance under obfuscation as a possible strategy for detecting dataset contamination and highlighting potential safety risks and interpretability issues for automated software systems.
\end{abstract}


\section{Introduction}

LLMs have become a key tool for automated software engineering. They are being used in a variety of contexts, with increasing autonomy \cite{hou2024largelanguagemodelssoftware}. Importantly, they are commonly used for code generation tasks, and their outputs are being integrated into software \cite{gao2024currentchallengessoftwareengineering}. These decisions are being made with expectations of performance, informed by evaluation. It is more important than ever to be able to fairly evaluate LLMs' use for code generation and problem-solving, and better understand the risks of the unintended behaviours they may exhibit, especially in the presence of dataset contamination.

Numerous studies have evaluated the performance of LLMs on standard datasets - there are even leaderboards ranking the performance of various models on them \cite{xia2025leetcodedatasettemporaldatasetrobust, mathleaderboard}. LLMs were also compared against humans, with studies suggesting better performance \cite{Coignion_2024}. However, the validity of these rankings is doubtful because the behaviour of the models is not guaranteed to be genuine. Researchers have raised concerns about the possibility of results reflecting dataset contamination and models' overfitting on training data \cite{xia2025leetcodedatasettemporaldatasetrobust}. These concerns challenge the validity and rigour of current evaluation approaches and their resulting conclusions. These limitations can, in turn, cause overestimation of LLMs' capabilities, unsafe deployment in production environments, and an altered public perception of the technology that can threaten its general adoption.

\begin{figure}[H]
\footnotesize
\begin{center}
\begin{minipage}{0.96\linewidth}
\begin{verbatim}
Write Python code to solve the following problem:

hJFFm two a8Gh4E se5Zuz GIJe! and mj<Cq of q&Xs m
and n %FE)2Xt(G#Oy, t$5 tDh the k$rJQH of the two
EPE^@W xGeX %E. The (gSFq<: run F(K@ DkN(;ss9r7 W
Bij>v be O (log (m + n) ). K@jb$T = = n 0 <= m <= 
w000 0 <= n <= w000 1 <= m + n <= 1000 - 106 <= 
GHnZ@ [i ], jk,e@ [i] <= 106
\end{verbatim}
\end{minipage}
\end{center}
\caption{Example of an obfuscated task.}
\label{fig:motivation}
\end{figure}

This paper contributes to this discussion by studying the LLMs behaviour when solving problems under obfuscated inputs. We perform an exhaustive evaluation, adding varying levels of noise to common benchmark tasks, up to a point where they are no longer recognisable. These obfuscated tasks were supplied to LLMs, and their outputs were evaluated. Figure~\ref{fig:motivation} shows an example of an obfuscated task - the text is unintelligible, and humans will not be able to understand the question or write a solution. However, we found that when this was asked of several LLMs, they were consistently able to solve this task, ``correctly'' identifying it as a problem of finding the median value in two sorted arrays, found in the LeetCode benchmark dataset. We show that this behaviour is more evident in tasks published before the models’ knowledge cutoff date, suggesting strong memorisation or overfitting to training data, rather than legitimate reasoning about the presented problem. We show that this limitation can be exploited, highlighting implications for safety. Based on the empirical results, we discuss the implications for benchmarking and evaluations of model behaviour, arguing for caution when designing experiments using standard datasets. We also propose measuring the decay of performance under obfuscation as a possible strategy for detecting dataset contamination. 
\begin{itemize}
    \item We empirically show that performance decay under extreme obfuscation is a practical indicator of dataset contamination and overfitting.
    \item We provide a quantitative analysis of performance decay across two standard evaluation datasets (LeetCode, MATH), revealing stark differences between new and contaminated tasks.
    \item We introduce the concept of eager pattern matching to describe the behaviour where LLMs solve tasks obfuscated beyond human recognition by exploiting spurious patterns, leading to incorrect solutions on new problems.
    \item We propose a reproducible framework for generating obfuscated tasks and testing dataset contamination.
    \item We discuss the broader implications for LLM evaluation practices, deployment in automated software systems, and risks arising from eager pattern matching and contamination.
\end{itemize}

Section~\ref{sec:background} introduces the background of this study and Section~\ref{sec:methodology} its methodology. Section~\ref{sec:results} presents the results, and Sections~\ref{sec:discussion}~and~\ref{sec:conclusions} discuss our findings and their implications.

\section{Background and Related Work}
\label{sec:background}

\subsection{Dataset contamination detection}

There exist benchmark datasets which have been extensively used in evaluating LLMs. They have become key resources, resulting in the construction of performance leaderboards \cite{mathleaderboard}, and gathering entire communities of researchers \cite{koch2021reducedreusedrecycledlife}. Important examples include the LeetCode \cite{nyandoggo_leetcode} and MATH \cite{mathdataset} datasets.

However, due to the relative importance of these datasets, they have also been extensively used in model training, which unfortunately corrupts measurements of performance and other original aims of the datasets \cite{roberts2023datacontaminationlenstime}. Significant efforts are being undertaken to combat these issues, most notably through temporal dataset splits \cite{xia2025leetcodedatasettemporaldatasetrobust} and dynamic dataset creation \cite{kiela2021dynabenchrethinkingbenchmarkingnlp}.

Methods of detecting LLM contamination have been proposed, such as token probability, completion overlap, and performance \cite{samuel2024datacontaminationdetectionmodern, ravaut2025comprehensivesurveycontaminationdetection}. Some of these methods have been shown to achieve high accuracies, but due to a lack of ground truth information and probabilistic LLM outputs, no method is perfectly reliable, and researchers are constantly looking to find new ways of testing contamination \cite{ravaut2025comprehensivesurveycontaminationdetection}. We contribute to this area of research by considering LLM behaviour and performance on obfuscated tasks as a method for detecting dataset contaminaton.

\subsection{LLM resilience to noise}

There have been multiple previous studies investigating the performance of LLMs on noisy or obfuscated tasks, primarily focusing on their resilience to input noise. Researchers see this as a topic of particular interest in the context of LLMs' interactions with humans, where noise is likely to be introduced, and might cause negative effects.

Wang et al. \cite{wang2024resiliencelargelanguagemodels} investigate the resilience of LLMs to five types of input noise, on tasks from the MMLU dataset \cite{mmluhendrycks2021measuringmassivemultitasklanguage}. Khandalkar et al. \cite{khandalkar2025impactnoisellmmodelsperformance} tested the performance of LLMs when noise is added to the ARC dataset \cite{arcclark2018thinksolvedquestionanswering}. Both studies report performance degradation across all models considered when noise is introduced and aim to improve robustness. Crucially, the rate of added noise was limited, with both papers only introducing noise levels equivalent up to about 10-35\% of the range we examine.

In contrast with prior studies, which have focused on improving LLM resilience to noise, we explicitly use performance decay under extreme obfuscation as a symptom of dataset contamination and overfitting. This highlights a gap in the literature surrounding evaluations of LLMs under extreme conditions and their consequences. To our knowledge, the angle of using performance decay analysis as a contamination detection strategy has not been explored in previous work. 

\section{Methodology}
\label{sec:methodology}

We conducted an experiment to examine the overperformance of LLMs on obfuscated tasks, systematically attempting to reproduce this phenomenon across a range of tasks and large language models, recording outcomes and intermediate stages.

\subsection{Datasets}

The dataset designated by us as LeetCode (Old) or OldLC was downloaded from HuggingFace \cite{lhoest2021datasetscommunitylibrarynatural, nyandoggo_leetcode}. We selected the first 20 questions from the dataset to be included in our experiment. The questions were originally sourced from the LeetCode platform \cite{leetcode_problemset}.

The LeetCode (New) or NewLC dataset was compiled by us, using the 20 most recent LeetCode questions at the time \cite{leetcode_problemset}. All questions were published in March 2025. The dataset comprises publicly available tasks and metadata accessed from the LeetCode Problemset \cite{leetcode_problemset}.

We also included a dataset of non-coding tasks. The MATH dataset, released in 2021, is one of the most well-known mathematical problem-solving benchmarks \cite{mathdataset}. It is used very frequently in LLM evaluations \cite{mathleaderboard}. The dataset was downloaded from HuggingFace \cite{lhoest2021datasetscommunitylibrarynatural, math_augmented_dataset}. We selected the first 20 questions from the dataset to be included in our experiment.

\subsection{Obfuscation methods}
\label{subsec:obfuscation_methods}

We considered multiple methods for obfuscating tasks through text augmentation. We identified the open-source \verb|nlpaug| library as the best starting point, given the multiple implemented methods for textual augmentation, designed to be used in machine learning experiments. We selected \textit{Typos}, which simulates human typing errors, and \textit{Deletions}, which randomly removes a proportion of words. As this functionality was not available in the \verb|nlpaug| library, we implemented \textit{Truncation} as a simple function removing text beyond a given point.

We use each obfuscation method to create 10 new versions of each task, obfuscating the text with increasing rates of augmentation. Figure~\ref{fig:obfuscation_examples} shows selected examples of obfuscated tasks.

\begin{figure}[H]
\footnotesize
\begin{center}
\begin{minipage}{0.82\linewidth}
\begin{verbatim}Given a signed 32-bit integ\end{verbatim}
\vspace{-0.5em}
A coding task obfuscated with Truncation (0.9 aug. rate)

\vspace{0.5em}
\begin{verbatim}Are given height. vertical lines (,)(,[]).
Find -, contains. can. that not container.\end{verbatim}
\vspace{-0.5em}
A coding task obfuscated with Deletion (0.7 aug. rate)

\vspace{0.5em}
\begin{verbatim}wplv$ \ [\ Xqrh {1 + \ xqrh {2 + \ QqTt 
{x} }} = \ sar5 [3] {1 + \ s@rf {x} }. \ ]\end{verbatim}
\vspace{-0.5em}
A math task obfuscated with Typos (0.5 aug. rate)

\vspace{0.5em}
\begin{verbatim}$ $ \ [| ^ + | = | (+) |. \] $ | + |. $\end{verbatim}
\vspace{-0.5em}
A math task obfuscated with Deletion (0.5 aug. rate)

\vspace{0.5em}
\begin{verbatim}Is $f(x) = 3^{x^2-3} \end{verbatim}
\vspace{-0.5em}
A math task obfuscated with Truncation (0.8 aug. rate)

\end{minipage}
\end{center}
\caption{Examples of obfuscated tasks.}
\label{fig:obfuscation_examples}
\end{figure}

\subsection{Large Language Models}

The models were chosen to come from a diverse range of developers, sizes, and inference costs. To ensure our evaluations of the NewLC datasets are not affected by contamination, we only considered models with knowledge cutoffs and release dates before March 2025. The set chosen represents the relative leaders of the industry \cite{fradkin2025, openrouter}.

\begin{itemize}
    \item Claude 3.5 Haiku, Anthropic, April 2024 \cite{anthropic2024claude3addendum})
    \item DeepSeek V3 (DeepSeek, December 2024 \cite{deepseek_v3_cite})
    \item Gemini 2.0 Flash (Google, February 2025 \cite{google_gemini20})
    \item Llama 3.3 (Meta, December 2024 \cite{meta_llama_2024})
    \item GPT-4o-mini (OpenAI, July 2024 \cite{openai_gpt4o}
\end{itemize}

The experiment used the OpenRouter platform \cite{openrouter} to send API requests to the LLM providers. The LLMs were prompted with the obfuscated task, along with brief instructions asking them to provide either Python code that solves this question or mathematical reasoning with an indicated final answer. In cases where the response was not in the requested format, they were re-prompted up to 3 times. If an LLM did not respond with a parsable response after that, it was marked as unable to answer that question correctly.

\subsection{Metric}

For coding tasks, input-output pairs present after the task were designated as test cases, usually 2-3 per question. To be considered correct, the LLM solution had to pass all test cases in automated evaluation. As LLM-generated code is generally considered unsafe \cite{codesafety}, we evaluated it in a sandboxed environment, blocking a predefined set of potentially malicious keywords and imports \cite{mertens2023dangerous}. We did not record any instances where code was blocked from executing.

For mathematical tasks, only the final answer was compared. To be considered correct, the LLM answer string had to match the original solution exactly, with minor formatting differences allowed.

\subsection{Implementation}

\begin{figure}[H]
    \centering
    \usetikzlibrary{positioning}
    \makebox[0.57\textwidth][c]{
    \begin{tikzpicture}[
        node distance=0.75cm,
        auto,
        thick,
        box/.style={draw, minimum width=2cm, minimum height=1.5cm, align=center}
    ]
        \node (dataset) [box] {Task};
        \node (datasetdesc) [right=0.25cm of dataset, align=left, text width=4cm] {\small From datasets: \begin{itemize}[left=0pt]
            \item OldLC
            \item NewLC
            \item MATH
        \end{itemize}};
        
        \node (obfuscate) [box, below=of dataset] {Obfuscate};
        \node (methods) [right=0.25cm of obfuscate, align=left, text width=4cm] {\small Methods: \begin{itemize}[left=0pt]
            \item Truncation
            \item Typos
            \item Deletion
        \end{itemize}};
        \node (severity) [right=2.25cm of obfuscate, align=left, text width=4cm] {\small Severity: \begin{itemize}[left=0pt]
            \item 1 not obfuscated
            \item 10 levels of \\augmentation
        \end{itemize}};

        \node (query) [box, below=of obfuscate] {LLM};
        \node (querydesc) [right=0.25cm of query, align=left, text width=3cm] {\small Models: \begin{itemize}[left=0pt]
            \item Claude 3.5 Haiku
            \item DeepSeek V3
            \item Gemini 2.0 Flash
            \item Llama 3.3
            \item GPT-4o-mini
        \end{itemize}};

        \node (eval) [box, below=of query] {Evaluate};
        \node (evaldesc) [right=0.25cm of eval, align=left, text width=3cm] {\small};

        \draw[->] (dataset) -- (obfuscate);
        \draw[->] (obfuscate) -- (query);
        \draw[->] (query) -- (eval);
    \end{tikzpicture}}
    \vspace{1em}
    \caption{Simplified experiment pipeline.}
    \label{fig:experiment_pipeline}
\end{figure}
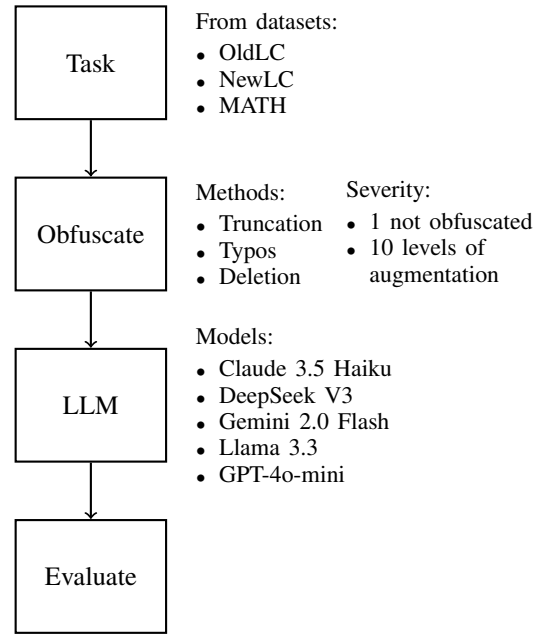

The software needed to implement and carry out the experiments is published in a GitHub repository associated with this project: \url{https://github.com/radzim/obfuscated}. An overview of the experiment pipeline is presented in Figure~\ref{fig:experiment_pipeline}.

\subsection{Human baselines}

In order to establish a baseline for what the expected behaviour should be, we asked 4 researchers with a high level of coding and math skills to report what percentage of tasks they believed to able to understand under different obfuscation methods and levels, by showing them randomly selected examples.

This is not the same metric as used in the LLM evaluation, as repeating the same experiment is impractical due to time constraints and memorisation. It is not directly comparable to actual performance, however, it provides a useful comparison baseline.

\subsection{Adversarial tasks}
\label{subsec:adversarial_tasks_method}

In order to further test our hypotheses, we manually crafted tasks that closely resemble well-known questions from the LeetCode dataset, but are in fact asking very different questions. Considering the previously mentioned ``median of two arrays'' problem, we present an adversarial example in Figure~\ref{fig:adversarial_task}.

\begin{figure}[H]
\footnotesize
\begin{center}
\begin{minipage}{0.96\linewidth}
\begin{verbatim}
Write Python code to solve the following problem:

Given two arrays nums1 and nums2 of size m and n 
respectively, return the medians of  the two 
arrays. The overall run time complexity should be.
Constraints:
0 <= m <= 1000
0 <= n <= 1000
1 <= m + n <= 2000
-106 <= nums1[i], nums[2] <= 106
\end{verbatim}
\end{minipage}
\end{center}
\caption{Adversarial task resembling the structure of the ``median of two sorted arrays'' task.}
\label{fig:adversarial_task}
\end{figure}

This problem is functionally different from the original in two key ways: it makes no mention of the inputs being sorted, and it asks for two separate medians of the two arrays, not one common. We will be asking the LLMs to solve it, observing if they are more likely to reason about it correctly, or rather eagerly pattern match to a similarly-looking known task.

\section{Results}
\label{sec:results}

We successfully conducted the experiment, creating 600 new tasks through augmentation to be used along the 60 original ones, eliciting 3300 LLM responses, and generating 3300 task evaluation scores. These intermediate artefacts are made available in the project's GitHub repository: \url{https://github.com/radzim/obfuscated}.

We report aggregate performance metrics for each combination of dataset, model, and augmentation rate. We report average scores achieved over the 20 tasks in each dataset and the 3 obfuscation methods considered. More detailed results reporting scores separated into the 3 obfuscation methods are presented in Appendices A and B.

\subsection{LeetCode datasets}

The overall result we were expecting in this part of our setup was accuracy decaying with augmentation, eventually reducing to zero when key details become obfuscated. This is consistent with the behaviour we see in the NewLC dataset, with a 49\% performance decay at 0.3 augmentation, and a 100\% decay at 0.8 and above.

\begin{figure}[H]
    \centering
    \includegraphics[width=0.42\textwidth]{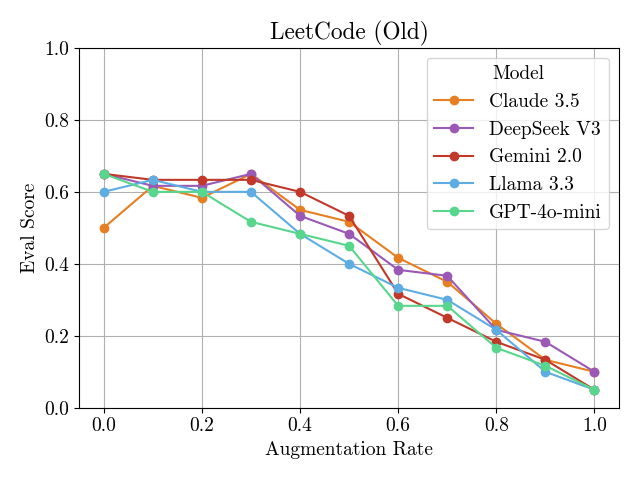}
    \includegraphics[width=0.42\textwidth]{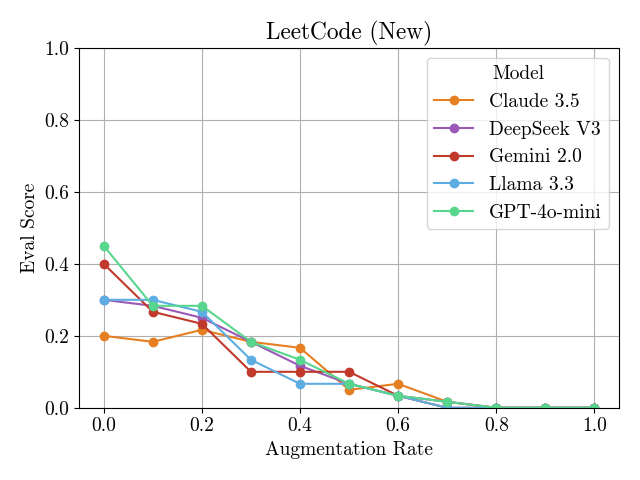}
    \caption{The performance of LLMs on the two LeetCode datasets, averaged across the three obfuscation methods. Performance across the augmentation methods is presented in Appendix A.}
    \label{fig:leetcode_performance}
\end{figure}

In contrast, the performance on the OldLC dataset did not suffer with increasing augmentation as much as expected. The performance on 0.3-augmented tasks was only 5\% lower than the original, and even some 1.0-augmented tasks were still being correctly solved. The performance on both LeetCode datasets is illustrated in Figure~\ref{fig:leetcode_performance}.

For coding tasks such as this, in our human baseline, tasks obfuscated with an augmentation rate above 0.5 were judged to be impossible to solve by any of the participants.
We believe many of the solutions presented by LLMs are not legitimate, but rather rely on recognising patterns and responding with solutions to previously seen problems. Some interesting examples of particularly impossible-looking solutions are highlighted in the next section. 

The accuracy on the NewLC dataset is much lower across all augmentation rates, and while the LLMs are not able to overcome the most extreme levels of obfuscation, they show very good error-correcting capabilities. We hypothesised whether the OldLC solutions could possibly be due to legitimate error-correcting capabilities of LLMs. We evaluated this using adversarial examples, and showed that gravitation towards previously seen tasks dominates error correction. Details of this are included in Appendix D.

We compare the relative performance decay of LLMs across the two LeetCode datasets. These contain very similar tasks, coming from the same source, but differ substantially in time of release (2015 vs 2025). The OldLC dataset is almost certainly in the training set of all of the LLMs examined, and it has been a key benchmark for years \cite{xia2025leetcodedatasettemporaldatasetrobust}. The New LeetCode dataset contains questions released after the release dates of all of the LLMs, which makes it unlikely for any of the questions to have been included in the training set.

\begin{figure}[H]
    \centering
    \includegraphics[width=0.42\textwidth]{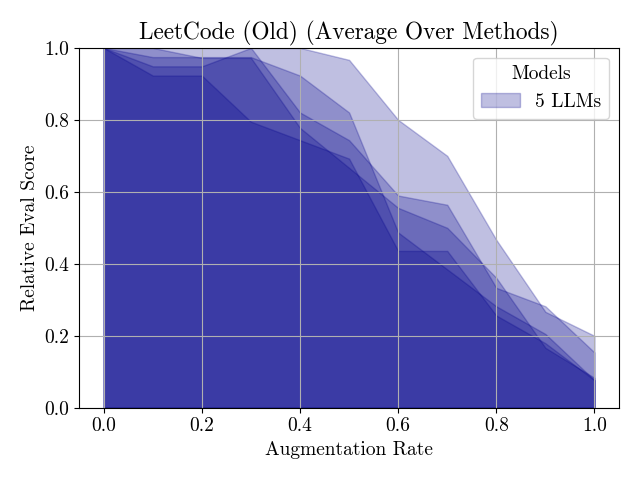}
    \includegraphics[width=0.42\textwidth]{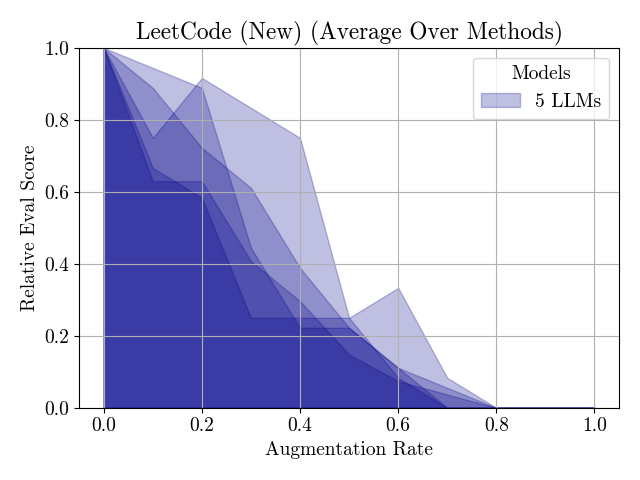}
    \caption{Comparison of the average performance decays of LLMs on the two Leetcode datasets.}
    \label{fig:leetcode_decay}
\end{figure}

Based on the comparison of the OldLC and NewLC datasets shown in Figure~\ref{fig:leetcode_decay}, we hypothesise that a slower rate of decay can be used as a sign of overtraining or inclusion in training data. The comparison between the decay on OldLC and NewLC highlights the stark difference in model behaviour under increasing augmention. Despite 900 attempts, not a single LLM had solved any task obfuscated at above 0.7 in the New dataset, while this has been achieved on the Old dataset in 13.6\% of cases - 122 times.

\subsection{MATH dataset}

The MATH dataset pre-dates the release of the LLMs and their knowledge cutoffs. Therefore, we are unable to compare the relative performance losses between questions included in the training sets and new ones. We evaluate the performance of LLMs on obfuscated tasks from the MATH dataset to examine if the behaviour we identified above in the OldLC dataset is present in other types of tasks.

According to the established human baselines, the highest augmentation rate where any of the researchers believed to be able to understand some questions was 0.4 for this dataset.

\begin{figure}[H]
    \centering
    \includegraphics[width=0.42\textwidth]{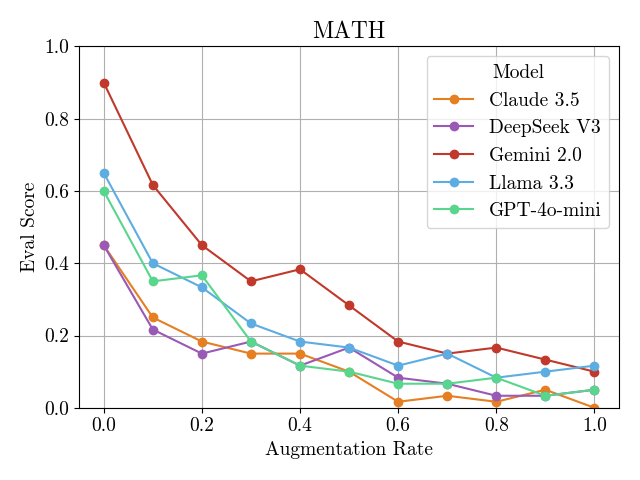}
    \caption{The average performance of selected LLMs on the 20 tasks in the dataset, averaged across the three obfuscation methods.}
    \label{fig:math_performance}
\end{figure}

Our evaluation presented in Figure~\ref{fig:math_performance} shows that a substantial number of questions are being solved far beyond that point, with many correct solutions even at the highest levels of obfuscation.

\begin{figure}[H]
    \centering
    \includegraphics[width=0.42\textwidth]{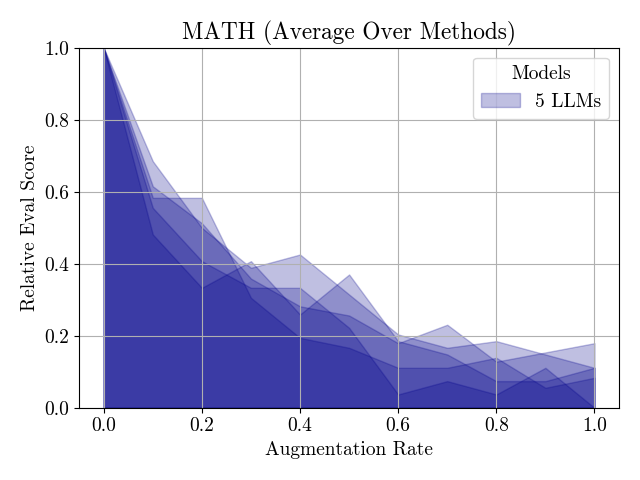}
    \includegraphics[width=0.42\textwidth]{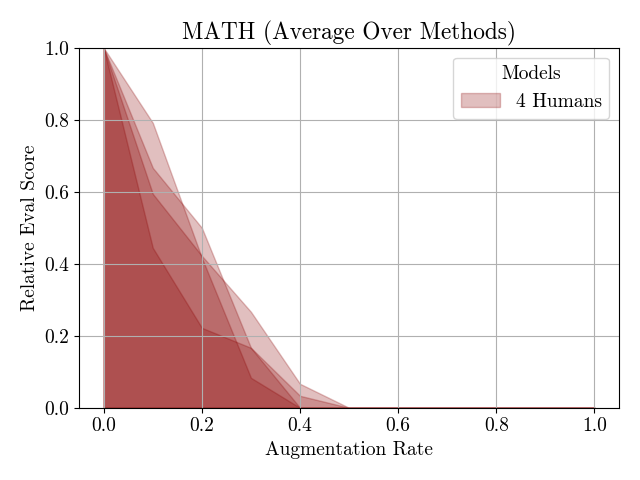}
    \caption{Comparison of the average performance decays of LLMs (evaluated) and humans (self-reported) on the MATH dataset. Performance under specific obfuscation methods is detailed in Appendix B.}
    \label{fig:math_decay}
\end{figure}

In Figure~\ref{fig:math_decay}, showing performance decay, we do see a steep initial drop, which we didn't see in the case of the OldLC dataset. This could suggest a lower degree of overtraining or perhaps not being included in the training set at all. However, the good performance on highly obfuscated tasks disproves that possibility. We hypothesise that the length of questions has key impact here, as MATH tasks on average were 4-6 times shorter than LeetCode tasks. Longer strings of information are known to be significantly more resilient to noise through language redundancy \cite{shannon1948mathematical}.

\subsection{Statistical testing}

Table~\ref{fig:leetcode_decay_metrics} presents some statistical properties of performance decay seen by our experiments. We measured the average augmentation level required to cause a 50\% and a 100\% drop in model performance, a linear model fit for the gradient of the decay function, and the average performance decay across all augmentation levels. Confidence interval width was calculated using parametric resampling, modelling the results for each augmentation level for each model as coming from a binomial distribution.

\begin{table}[H]
\centering
\begin{tabular}{lccc}
\toprule
Metric & OldLC & NewLC & MATH \\
\midrule
50\% decay & 0.70 ± 0.08 & 0.29 ± 0.06 & 0.19 ± 0.03 \\
100\% decay & 1.00 ± 0.00 & 0.70 ± 0.05 & 1.00 ± 0.00 \\
Gradient & -0.98 ± 0.05 & -1.42 ± 0.14 & -0.73 ± 0.06 \\
Average & 0.67 ± 0.04 & 0.34 ± 0.04 & 0.33 ± 0.03 \\
\bottomrule
\end{tabular}
\caption{Comparison of several decay metrics on the three datasets.}
\label{fig:leetcode_decay_metrics}
\end{table}

We highlight the 50\% decay statistic as most useful at this stage, due to its simple definition and calculation, large contrasts, and relatively low variance.

Examining the confidence intervals for the two LeetCode datasets, we find that the differences in LLM behaviour between these two similar datasets are significant, with OldLC decaying slower in all four metrics considered.

\subsection{Illustrative Examples}

As shown in Figure~\ref{fig:obfuscation_examples_solved}, each of the examples originally presented in Subsection~\ref{subsec:obfuscation_methods} was solved by at least one of the LLMs, despite not containing sufficient information to be solved using legitimate reasoning.

\begin{figure}[H]
\footnotesize
\begin{center}
\begin{minipage}{0.82\linewidth}
\begin{verbatim}Given a signed 32-bit integ\end{verbatim}
\vspace{-0.5em}
``Reverse Integer''- solved by all 5 LLMs evaluated

\vspace{0.5em}
\begin{verbatim}Are given height. vertical lines (,)(,[]).
Find -, contains. can. that not container.\end{verbatim}
\vspace{-0.5em}
``Container With Most Water'' -  solved by Claude 3.5, DeepSeek~V3, and Gemini 2.0

\vspace{0.5em}
\begin{verbatim}wplv$ \ [\ Xqrh {1 + \ xqrh {2 + \ QqTt 
{x} }} = \ sar5 [3] {1 + \ s@rf {x} }. \ ]\end{verbatim}
\vspace{-0.5em}
Answer: \verb|49| - solved by DeepSeek V3 and Gemini 2.0

\vspace{0.5em}
\begin{verbatim}$ $ \ [| ^ + | = | (+) |. \] $ | + |. $\end{verbatim}
\vspace{-0.5em}
Answer: \verb|1| - solved by Gemini 2.0 and Llama 3.3

\vspace{0.5em}
\begin{verbatim}Is $f(x) = 3^{x^2-3} \end{verbatim}
\vspace{-0.5em}
Answer: \verb|even| - solved by Gemini 2.0

\end{minipage}
\end{center}
\caption{Examples of obfuscated tasks along with solutions.}
\label{fig:obfuscation_examples_solved}
\end{figure}

The LLMs were also tasked with the adversarial example we constructed in Subsection~\ref{subsec:adversarial_tasks_method}, asking them to find the two median values of two unsorted arrays. Surprisingly, all 5 LLMs queried replied with code for finding a single median of two sorted arrays - a clearly incorrect answer to the simple question, influenced by the stylistically similar task in its training data. This example illustrates how LLMs can be influenced by irrelevant patterns, causing them to respond to relatively simple tasks in unpredictable ways. 

\subsection{Performance of different LLMs}

We compared the initial performance and rate of decay across the 5 LLMs used in evaluation. This analysis is presented in Appendix C. While we observed some pairwise statistically significant differences, we found all 5 LLMs to exhibit the same characteristic behaviour and trends across all datasets and obfuscation methods considered.

\subsection{Characteristics of datasets}

The impact of obfuscation on the performance of LLMs on the 3 datasets was significantly different, with all 5 LLMs showing the quickest initial decline in MATH, followed by NewLC, and lastly OldLC.

It is difficult to compare effects between datasets, as beyond the issues of overtraining and contamination, the decay and overall behaviour could be affected by question style, length and information redundancy. However, as the two LeetCode datasets are very similar in all of the aforementioned aspects, we can make a meaningful comparison.

\section{Discussion}
\label{sec:discussion}

LLMs can solve tasks obfuscated beyond recognition and missing key details. This indicates that this is being done through memorisation and eager pattern matching rather than through genuine reasoning. Benchmarking models on contaminated datasets is therefore unreliable and likely overestimates real performance. We have shown that this behaviour exists on different types of tasks, in both coding and mathematical reasoning.

We showed that some correct solutions are not positive indicators of model capabilities, but rather artefacts of overtraining, casting doubt on the legitimacy of other solutions offered by the model. We add to the voices calling for dynamic benchmarks and controlled datasets which allow for real assessments of LLM behaviour and capabilities, and for key evaluation datasets to be accompanied with semantically similar yet unsolvable variations, to be used as a control sample against contamination.

The rates at which performance decays with task augmentation differ significantly between contaminated and new datasets. While it is not a new finding that many LLMs are severely overfit to training data and their performance can't be trusted on public benchmarks \cite{kiela2021dynabenchrethinkingbenchmarkingnlp}, the severity of this is yet to be understood, along with its impact on properties like model robustness or sensitivity to changes in user input.

LLM performance and resilience to obfuscation sharply decrease with obfuscation in unseen datasets, confirming contamination issues and invalidating their performance on previously seen data.

While this behaviour may appear to indicate strong performance, it is misleading and driven by overfitting and memorisation rather than true reasoning. In real-world applications, this eager pattern matching to familiar tasks can negatively impact performance on unseen problems. We show that this overcorrection towards training problems can cause unpredictable and opaque behaviour, introducing potential safety risks, particularly when LLMs are deployed in critical systems.

LLMs are increasingly deployed in multi-agent systems, where agents exchange messages with one another to achieve a goal. Humans must keep control of these systems, especially when they are critical. However, the findings of this study suggest that LLMs can operate with certain success, exchanging messages that humans cannot understand (i.e., obfuscated text), mirroring the famous example where two agents developed by Facebook in 2019 began communicating in their language \cite{lammin2021bobalice}. This lack of understanding affects how humans interpret these multi-agent systems, undermining human oversight and control. This issue is known as intellectual debt when we know that a system operates as expected but do not understand how~\cite{Zittrain_2022}. A lack of mechanisms to mitigate the intellectual debt threatens the sustainability and adoption of AI-based software systems~\cite{cabrera2024selfsustainingsoftwaresystemss4,cabrera2025systemsengineeringapproachtimes}.

\section{Conclusions}
\label{sec:conclusions}

This study examined the behaviour of LLMs on obfuscated tasks across three datasets, revealing strong indications that similar behaviour is likely present in other benchmarks. We observed clear differences in performance patterns across the datasets, suggesting the need to expand evaluations to additional datasets to generalise findings and explore dataset-specific effects.

Our analysis shows that performance decay under extreme obfuscation is closely tied to dataset contamination and overfitting, inferring strong relationships between inclusion in the training data and observed behaviour. We hypothesise that the decay statistics could be used as a measure of overtraining on a particular dataset and contamination detection. However, with just three sets of summary statistics, we are unable to account for confounding factors or test any such detection systems. In future research, it would be valuable to conduct a similar analysis for additional datasets, including those selectively used in training, to collect data needed to develop a robust dataset contamination detection system.

In conclusion, we demonstrate that LLM performance evaluations may be misleading, as they reflect memorisation and eager pattern matching rather than true reasoning. Our findings highlight the urgent need for rigorous evaluation practices that account for dataset contamination. We also underscore the potential unintended consequences of this model behaviour and urge caution when deploying LLMs in code generation systems. 


\begin{thebibliography}{10}
\providecommand{\url}[1]{#1}
\csname url@samestyle\endcsname
\providecommand{\newblock}{\relax}
\providecommand{\bibinfo}[2]{#2}
\providecommand{\BIBentrySTDinterwordspacing}{\spaceskip=0pt\relax}
\providecommand{\BIBentryALTinterwordstretchfactor}{4}
\providecommand{\BIBentryALTinterwordspacing}{\spaceskip=\fontdimen2\font plus
\BIBentryALTinterwordstretchfactor\fontdimen3\font minus \fontdimen4\font\relax}
\providecommand{\BIBforeignlanguage}[2]{{%
\expandafter\ifx\csname l@#1\endcsname\relax
\typeout{** WARNING: IEEEtran.bst: No hyphenation pattern has been}%
\typeout{** loaded for the language `#1'. Using the pattern for}%
\typeout{** the default language instead.}%
\else
\language=\csname l@#1\endcsname
\fi
#2}}
\providecommand{\BIBdecl}{\relax}
\BIBdecl

\bibitem{hou2024largelanguagemodelssoftware}
\BIBentryALTinterwordspacing
X.~Hou, Y.~Zhao, Y.~Liu, Z.~Yang, K.~Wang, L.~Li, X.~Luo, D.~Lo, J.~Grundy, and H.~Wang, ``Large language models for software engineering: A systematic literature review,'' 2024. [Online]. Available: \url{https://arxiv.org/abs/2308.10620}
\BIBentrySTDinterwordspacing

\bibitem{gao2024currentchallengessoftwareengineering}
\BIBentryALTinterwordspacing
C.~Gao, X.~Hu, S.~Gao, X.~Xia, and Z.~Jin, ``The current challenges of software engineering in the era of large language models,'' 2024. [Online]. Available: \url{https://arxiv.org/abs/2412.14554}
\BIBentrySTDinterwordspacing

\bibitem{xia2025leetcodedatasettemporaldatasetrobust}
\BIBentryALTinterwordspacing
Y.~Xia, W.~Shen, Y.~Wang, J.~K. Liu, H.~Sun, S.~Wu, J.~Hu, and X.~Xu, ``Leetcodedataset: A temporal dataset for robust evaluation and efficient training of code llms,'' 2025. [Online]. Available: \url{https://arxiv.org/abs/2504.14655}
\BIBentrySTDinterwordspacing

\bibitem{mathleaderboard}
``Math word problem solving on math leaderboard,'' \url{https://paperswithcode.com/sota/math-word-problem-solving-on-math}, accessed: 2025-05-23.

\bibitem{Coignion_2024}
\BIBentryALTinterwordspacing
T.~Coignion, C.~Quinton, and R.~Rouvoy, ``A performance study of llm-generated code on leetcode,'' in \emph{Proceedings of the 28th International Conference on Evaluation and Assessment in Software Engineering}, ser. EASE 2024.\hskip 1em plus 0.5em minus 0.4em\relax ACM, Jun. 2024, p. 79–89. [Online]. Available: \url{http://dx.doi.org/10.1145/3661167.3661221}
\BIBentrySTDinterwordspacing

\bibitem{koch2021reducedreusedrecycledlife}
\BIBentryALTinterwordspacing
B.~Koch, E.~Denton, A.~Hanna, and J.~G. Foster, ``Reduced, reused and recycled: The life of a dataset in machine learning research,'' 2021. [Online]. Available: \url{https://arxiv.org/abs/2112.01716}
\BIBentrySTDinterwordspacing

\bibitem{nyandoggo_leetcode}
NyanDoggo, ``Leetcode dataset,'' \url{https://huggingface.co/datasets/NyanDoggo/leetcode}, 2025, accessed: 2025-05-23.

\bibitem{mathdataset}
\BIBentryALTinterwordspacing
D.~Hendrycks, C.~Burns, S.~Kadavath, A.~Arora, S.~Basart, E.~Tang, D.~Song, and J.~Steinhardt, ``Measuring mathematical problem solving with the math dataset,'' 2021. [Online]. Available: \url{https://arxiv.org/abs/2103.03874}
\BIBentrySTDinterwordspacing

\bibitem{roberts2023datacontaminationlenstime}
\BIBentryALTinterwordspacing
M.~Roberts, H.~Thakur, C.~Herlihy, C.~White, and S.~Dooley, ``Data contamination through the lens of time,'' 2023. [Online]. Available: \url{https://arxiv.org/abs/2310.10628}
\BIBentrySTDinterwordspacing

\bibitem{kiela2021dynabenchrethinkingbenchmarkingnlp}
\BIBentryALTinterwordspacing
D.~Kiela, M.~Bartolo, Y.~Nie, D.~Kaushik, A.~Geiger, Z.~Wu, B.~Vidgen, G.~Prasad, A.~Singh, P.~Ringshia, Z.~Ma, T.~Thrush, S.~Riedel, Z.~Waseem, P.~Stenetorp, R.~Jia, M.~Bansal, C.~Potts, and A.~Williams, ``Dynabench: Rethinking benchmarking in nlp,'' 2021. [Online]. Available: \url{https://arxiv.org/abs/2104.14337}
\BIBentrySTDinterwordspacing

\bibitem{samuel2024datacontaminationdetectionmodern}
\BIBentryALTinterwordspacing
V.~Samuel, Y.~Zhou, and H.~P. Zou, ``Towards data contamination detection for modern large language models: Limitations, inconsistencies, and oracle challenges,'' 2024. [Online]. Available: \url{https://arxiv.org/abs/2409.09927}
\BIBentrySTDinterwordspacing

\bibitem{ravaut2025comprehensivesurveycontaminationdetection}
\BIBentryALTinterwordspacing
M.~Ravaut, B.~Ding, F.~Jiao, H.~Chen, X.~Li, R.~Zhao, C.~Qin, C.~Xiong, and S.~Joty, ``A comprehensive survey of contamination detection methods in large language models,'' 2025. [Online]. Available: \url{https://arxiv.org/abs/2404.00699}
\BIBentrySTDinterwordspacing

\bibitem{wang2024resiliencelargelanguagemodels}
\BIBentryALTinterwordspacing
B.~Wang, C.~Wei, Z.~Liu, G.~Lin, and N.~F. Chen, ``Resilience of large language models for noisy instructions,'' 2024. [Online]. Available: \url{https://arxiv.org/abs/2404.09754}
\BIBentrySTDinterwordspacing

\bibitem{mmluhendrycks2021measuringmassivemultitasklanguage}
\BIBentryALTinterwordspacing
D.~Hendrycks, C.~Burns, S.~Basart, A.~Zou, M.~Mazeika, D.~Song, and J.~Steinhardt, ``Measuring massive multitask language understanding,'' 2021. [Online]. Available: \url{https://arxiv.org/abs/2009.03300}
\BIBentrySTDinterwordspacing

\bibitem{khandalkar2025impactnoisellmmodelsperformance}
\BIBentryALTinterwordspacing
N.~Khandalkar, P.~Yadav, K.~Shinde, L.~B. Ramegowda, and R.~Das, ``Impact of noise on llm-models performance in abstraction and reasoning corpus (arc) tasks with model temperature considerations,'' 2025. [Online]. Available: \url{https://arxiv.org/abs/2504.15903}
\BIBentrySTDinterwordspacing

\bibitem{arcclark2018thinksolvedquestionanswering}
\BIBentryALTinterwordspacing
P.~Clark, I.~Cowhey, O.~Etzioni, T.~Khot, A.~Sabharwal, C.~Schoenick, and O.~Tafjord, ``Think you have solved question answering? try arc, the ai2 reasoning challenge,'' 2018. [Online]. Available: \url{https://arxiv.org/abs/1803.05457}
\BIBentrySTDinterwordspacing

\bibitem{lhoest2021datasetscommunitylibrarynatural}
\BIBentryALTinterwordspacing
Q.~Lhoest, A.~V. del Moral, Y.~Jernite, A.~Thakur, P.~von Platen, S.~Patil, J.~Chaumond, M.~Drame, J.~Plu, L.~Tunstall, J.~Davison, M.~Šaško, G.~Chhablani, B.~Malik, S.~Brandeis, T.~L. Scao, V.~Sanh, C.~Xu, N.~Patry, A.~McMillan-Major, P.~Schmid, S.~Gugger, C.~Delangue, T.~Matussière, L.~Debut, S.~Bekman, P.~Cistac, T.~Goehringer, V.~Mustar, F.~Lagunas, A.~M. Rush, and T.~Wolf, ``Datasets: A community library for natural language processing,'' 2021. [Online]. Available: \url{https://arxiv.org/abs/2109.02846}
\BIBentrySTDinterwordspacing

\bibitem{leetcode_problemset}
\BIBentryALTinterwordspacing
{LeetCode}, ``Leetcode problemset,'' 2025, accessed: 2025-05-25. [Online]. Available: \url{https://leetcode.com/problemset/}
\BIBentrySTDinterwordspacing

\bibitem{math_augmented_dataset}
nivektk, ``Math augmented dataset,'' \url{https://huggingface.co/datasets/nivektk/math-augmented-dataset}, 2021, accessed: 2025-05-27.

\bibitem{fradkin2025}
\BIBentryALTinterwordspacing
A.~Fradkin, ``Demand for llms: Descriptive evidence on substitution, market expansion, and multihoming,'' 2025. [Online]. Available: \url{https://arxiv.org/abs/2504.15440}
\BIBentrySTDinterwordspacing

\bibitem{openrouter}
{OpenRouter}, ``Openrouter: Unified api and playground for large language models,'' \url{https://openrouter.ai}, accessed: 2025-05-21.

\bibitem{anthropic2024claude3addendum}
\BIBentryALTinterwordspacing
Anthropic, ``Claude 3 model card october addendum,'' 2024, accessed: 2025-05-23. [Online]. Available: \url{https://assets.anthropic.com/m/1cd9d098ac3e6467/original/Claude-3-Model-Card-October-Addendum.pdf}
\BIBentrySTDinterwordspacing

\bibitem{deepseek_v3_cite}
\BIBentryALTinterwordspacing
{DeepSeek Inc.}, ``Introducing deepseek-v3,'' December 2024, accessed: 2025-05-28. [Online]. Available: \url{https://api-docs.deepseek.com/news/news1226}
\BIBentrySTDinterwordspacing

\bibitem{google_gemini20}
Google, ``Gemini 2.0 flash,'' \url{https://openrouter.ai/google/gemini-2.0-flash-001}, February 2025, accessed May 2025.

\bibitem{meta_llama_2024}
\BIBentryALTinterwordspacing
Meta, ``Llama 3.3 70b instruct,'' 2024, accessed: 2025-04-01. [Online]. Available: \url{https://huggingface.co/meta-llama/Llama-3.3-70B-Instruct}
\BIBentrySTDinterwordspacing

\bibitem{openai_gpt4o}
OpenAI, ``Gpt-4o-mini,'' \url{https://openrouter.ai/openai/gpt-4o-mini}, July 2024, accessed May 2025.

\bibitem{codesafety}
\BIBentryALTinterwordspacing
J.~Wang, X.~Luo, L.~Cao, H.~He, H.~Huang, J.~Xie, A.~Jatowt, and Y.~Cai, ``Is your ai-generated code really safe? evaluating large language models on secure code generation with codeseceval,'' 2024. [Online]. Available: \url{https://arxiv.org/abs/2407.02395}
\BIBentrySTDinterwordspacing

\bibitem{mertens2023dangerous}
X.~Mertens, ``Keeping an eye on dangerous python modules,'' \url{https://isc.sans.edu/diary/27514}, 2023, sANS Internet Storm Center.

\bibitem{shannon1948mathematical}
C.~E. Shannon, ``A mathematical theory of communication,'' \emph{Bell System Technical Journal}, vol.~27, no.~3, pp. 379--423, 1948.

\bibitem{lammin2021bobalice}
\BIBentryALTinterwordspacing
H.~Lammin, ``What are bob and alice saying? [mis]communication and intermediation between language and code,'' in \emph{Language Games}, L.~Aceti, S.~Calvert, and H.~Lammin, Eds.\hskip 1em plus 0.5em minus 0.4em\relax Cambridge, MA: LEA / MIT Press, 2021, published online: March 15, 2022. [Online]. Available: \url{https://api.semanticscholar.org/CorpusID:272648313}
\BIBentrySTDinterwordspacing

\bibitem{Zittrain_2022}
J.~Zittrain, \emph{Intellectual Debt: With Great Power Comes Great Ignorance}, ser. Cambridge Law Handbooks.\hskip 1em plus 0.5em minus 0.4em\relax Cambridge University Press, 2022, p. 176–184.

\bibitem{cabrera2024selfsustainingsoftwaresystemss4}
\BIBentryALTinterwordspacing
C.~Cabrera, A.~Paleyes, and N.~D. Lawrence, ``Self-sustaining software systems (s4): Towards improved interpretability and adaptation,'' in \emph{Proceedings of the 1st International Workshop on New Trends in Software Architecture}, ser. SATrends '24.\hskip 1em plus 0.5em minus 0.4em\relax New York, NY, USA: Association for Computing Machinery, 2024, p. 5–9. [Online]. Available: \url{https://doi.org/10.1145/3643657.3643910}
\BIBentrySTDinterwordspacing

\bibitem{cabrera2025systemsengineeringapproachtimes}
\BIBentryALTinterwordspacing
C.~Cabrera, V.~Bastidas, J.~Schooling, and N.~D. Lawrence, ``The systems engineering approach in times of large language models,'' in \emph{Proceedings of the 58th Hawaii International Conference on System Sciences}, 2025. [Online]. Available: \url{https://doi.org/10.1145/3643657.3643910}
\BIBentrySTDinterwordspacing

\end{thebibliography}

\appendices

\section{Performance by obfuscation method}

\begin{figure}[H]
    \centering
    \includegraphics[width=0.23\textwidth]{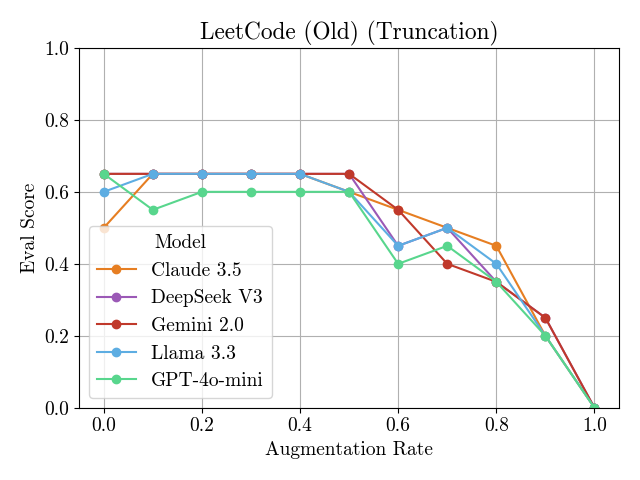}
    \includegraphics[width=0.23\textwidth]{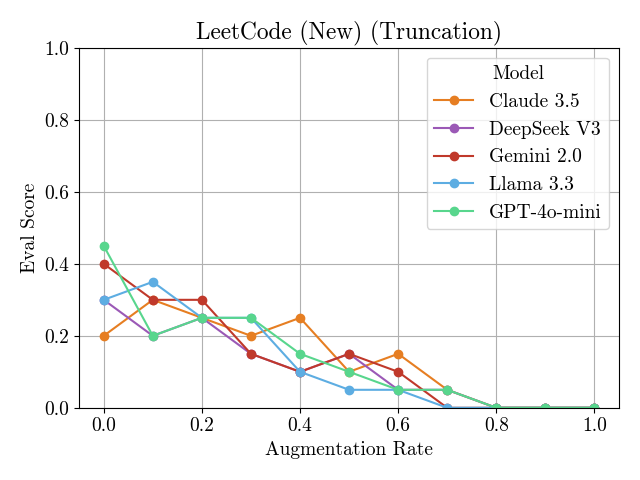}
    \includegraphics[width=0.23\textwidth]{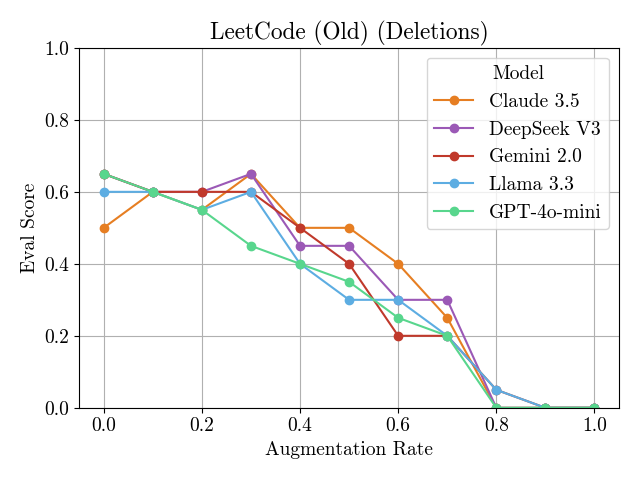}
    \includegraphics[width=0.23\textwidth]{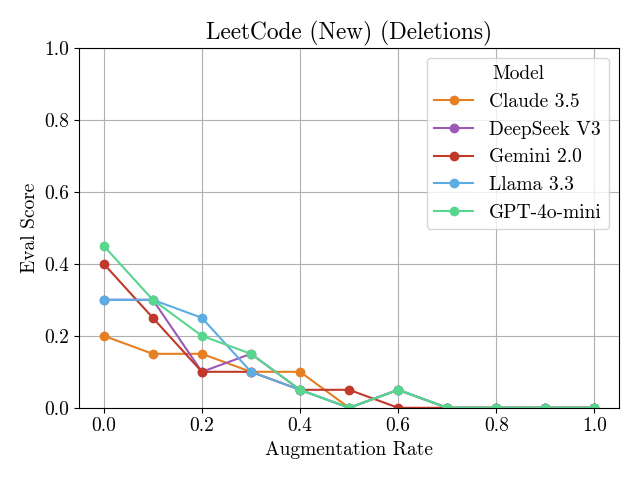}
    \includegraphics[width=0.23\textwidth]{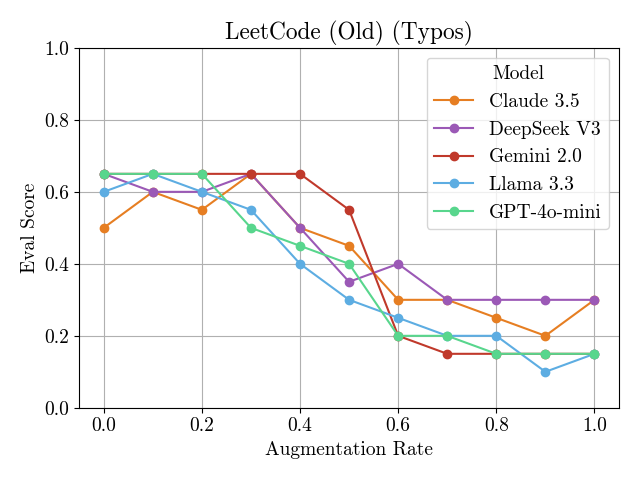}
    \includegraphics[width=0.23\textwidth]{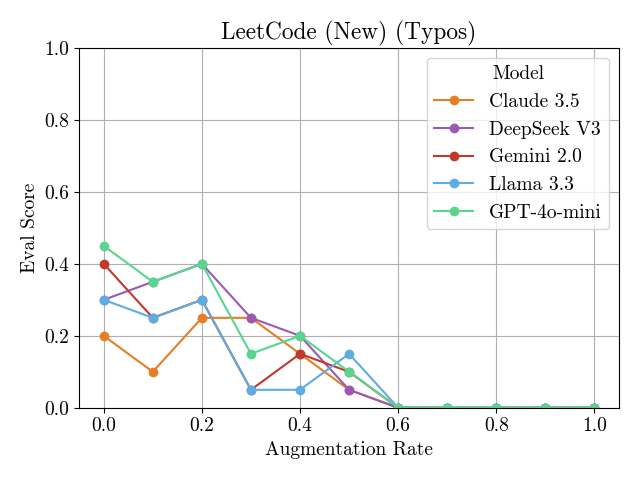}
    \caption{The average performance of selected LLMs on the 2 LeetCode datasets, across three obfuscation methods.}
    \label{fig:leetcode_performance_3_methods}
\end{figure}

\begin{figure}[H]
    \centering
    \includegraphics[width=0.23\textwidth]{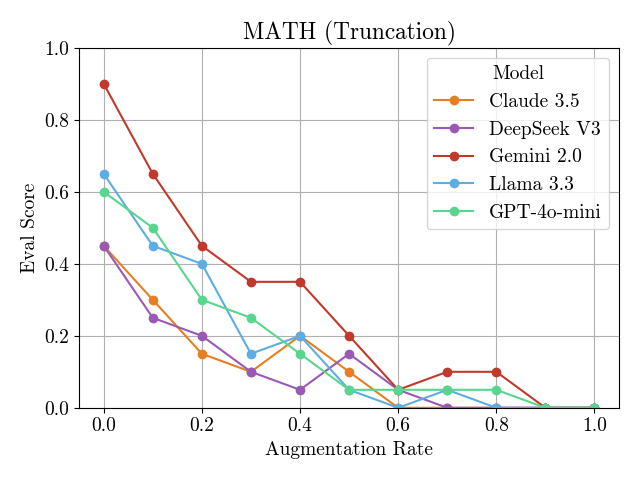}
    \\\includegraphics[width=0.23\textwidth]{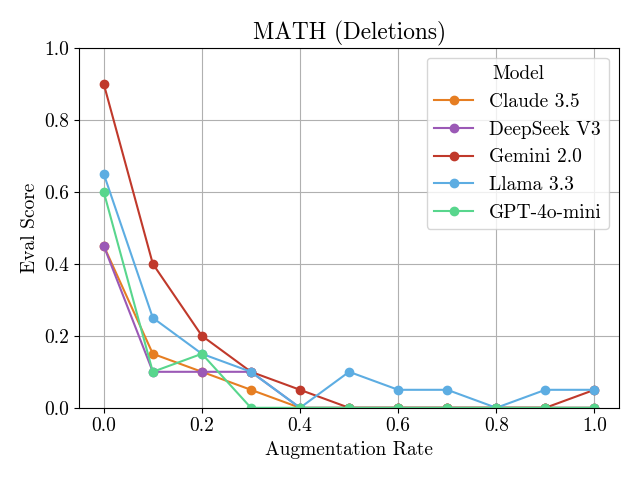}
    \\\includegraphics[width=0.23\textwidth]{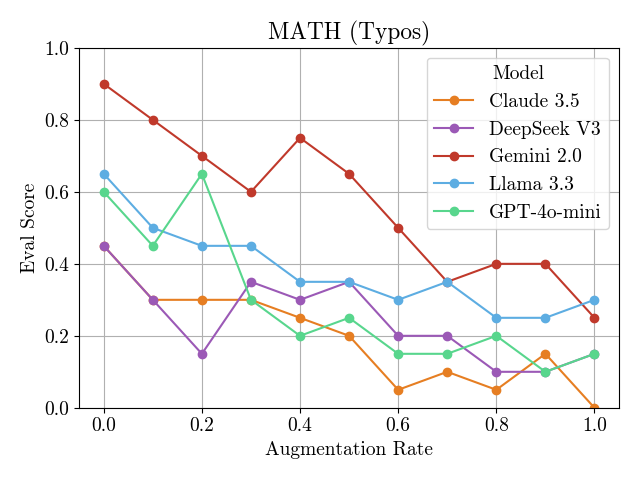}
    \caption{The average performance of selected LLMs on the 20 tasks in the MATH dataset, across three obfuscation methods: Truncation, Deletions, and Typos.}
    \label{fig:math_performance_3_methods}
\end{figure}

\section{Detailed performance decay}

\begin{figure}[H]
    \centering
    \includegraphics[width=0.23\textwidth]{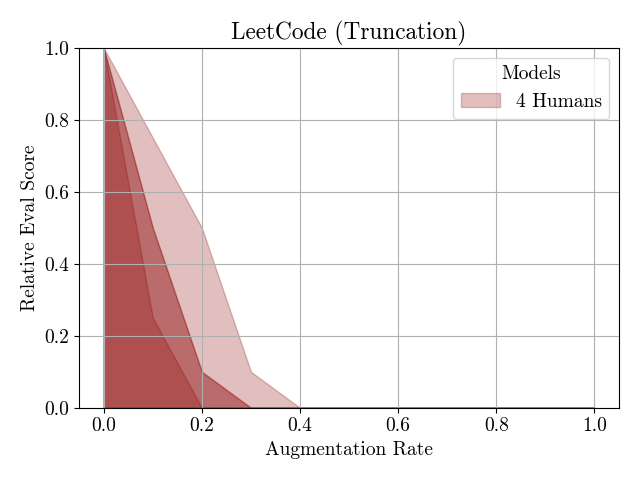}
    \includegraphics[width=0.23\textwidth]{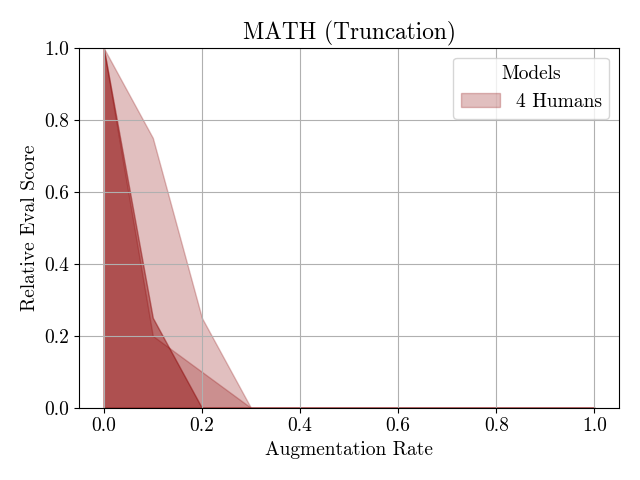}
    \includegraphics[width=0.23\textwidth]{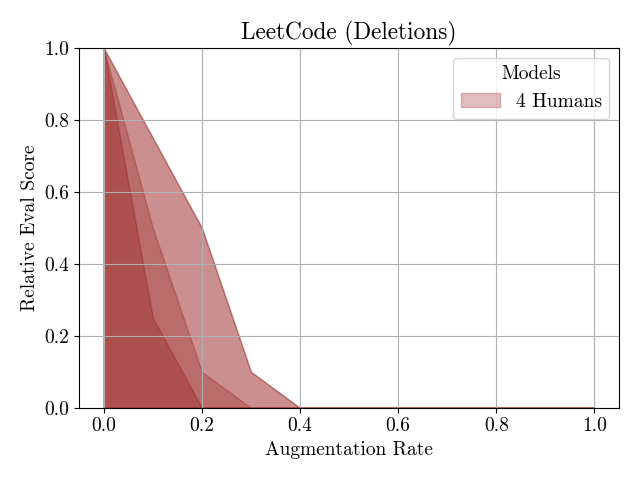}
    \includegraphics[width=0.23\textwidth]{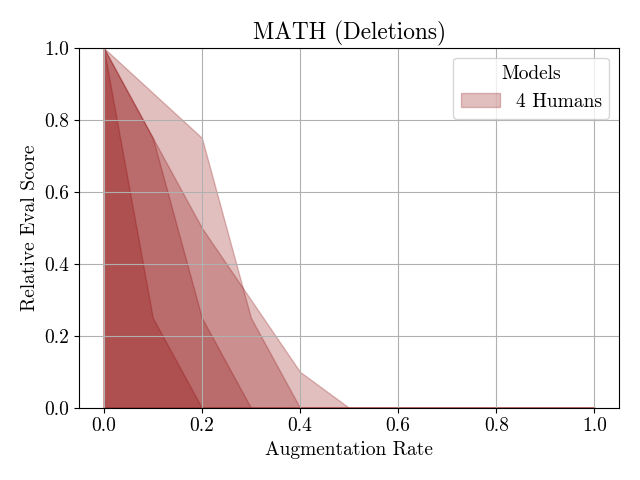}
    \includegraphics[width=0.23\textwidth]{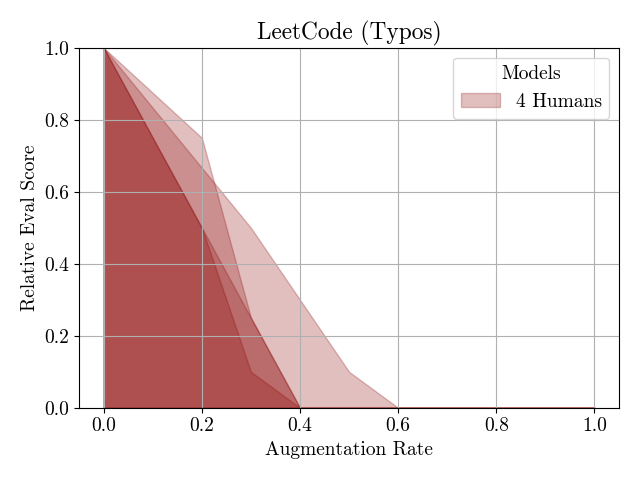}
    \includegraphics[width=0.23\textwidth]{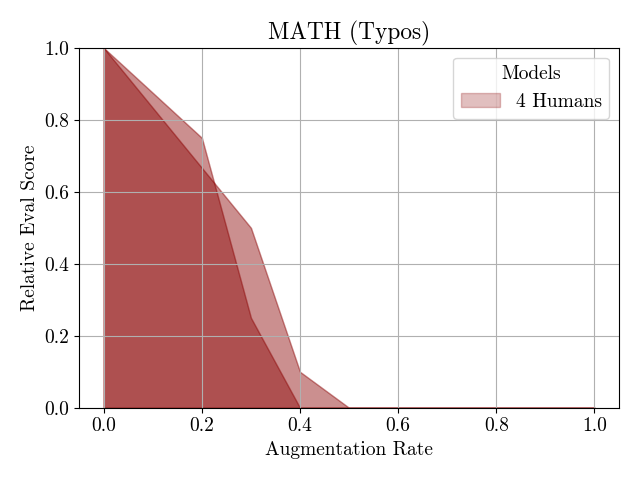}
    \caption{The self-reported average performance decay of humans on the datasets under different obfuscation methods.}
    \label{fig:human_decay_3_methods}
\end{figure}

\begin{figure}[H]
    \centering
    \includegraphics[width=0.23\textwidth]{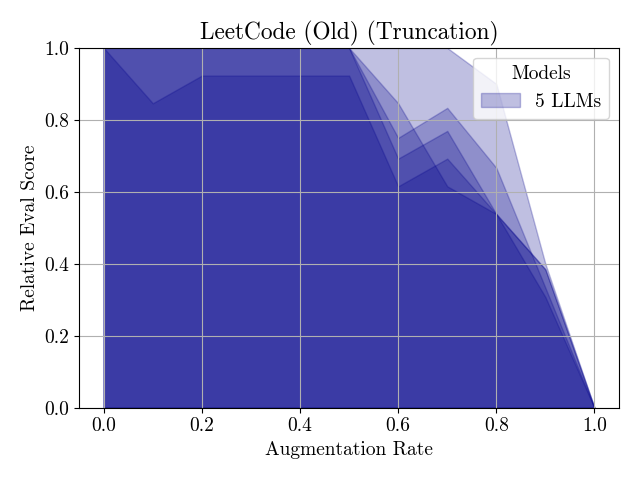}
    \includegraphics[width=0.23\textwidth]{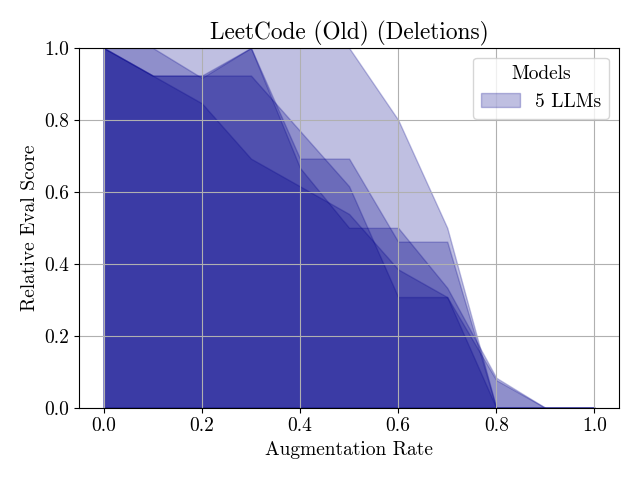}
    \includegraphics[width=0.23\textwidth]{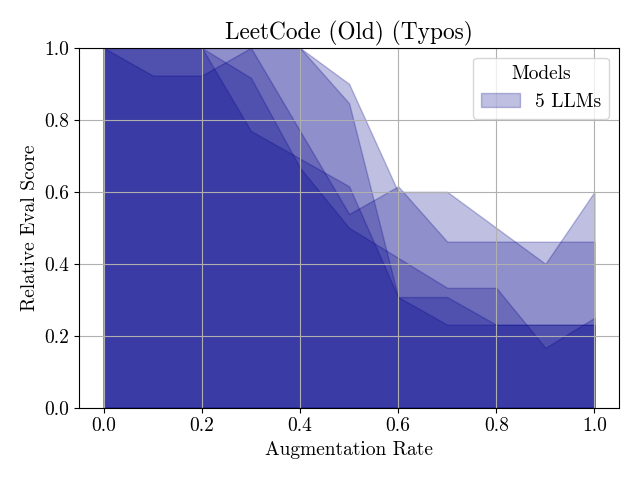}
    \includegraphics[width=0.23\textwidth]{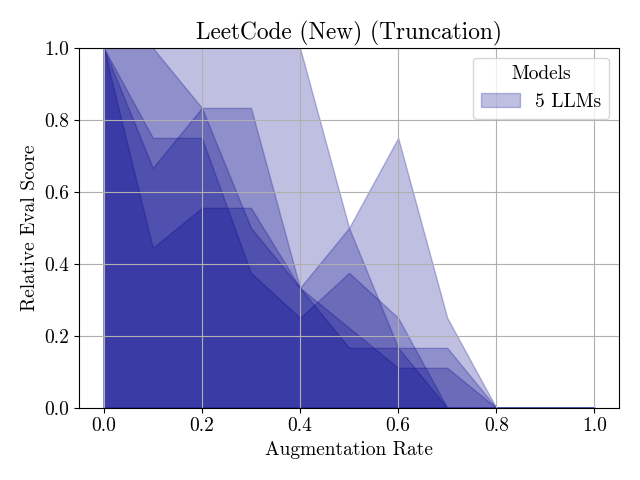}
    \includegraphics[width=0.23\textwidth]{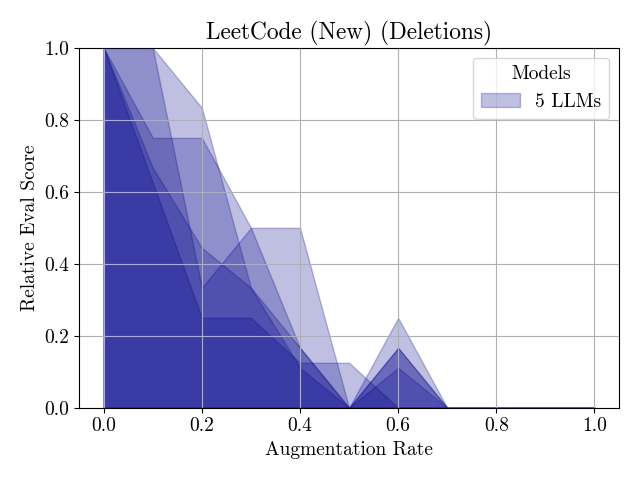}
    \includegraphics[width=0.23\textwidth]{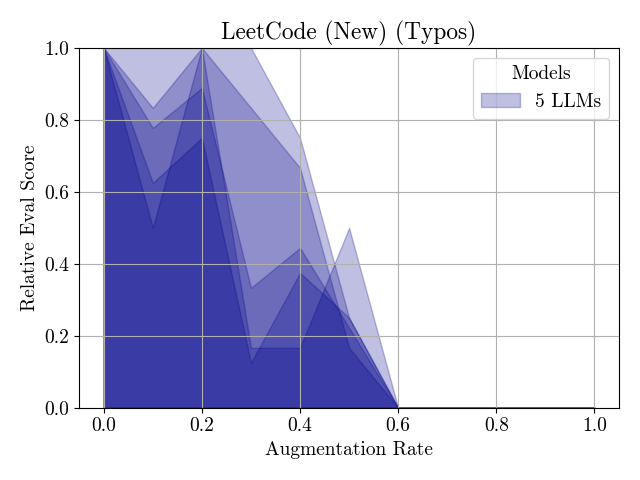}
    \caption{The average performance decay of LLMs on the LeetCode datasets under different obfuscation methods.}
    \label{fig:leetcode_decay_3_methods}
\end{figure}

\begin{figure}[H]
    \centering
    \includegraphics[width=0.23\textwidth]{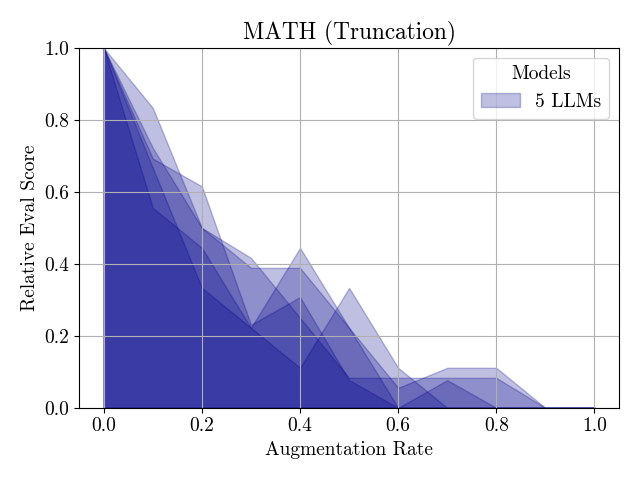}
    \\\includegraphics[width=0.23\textwidth]{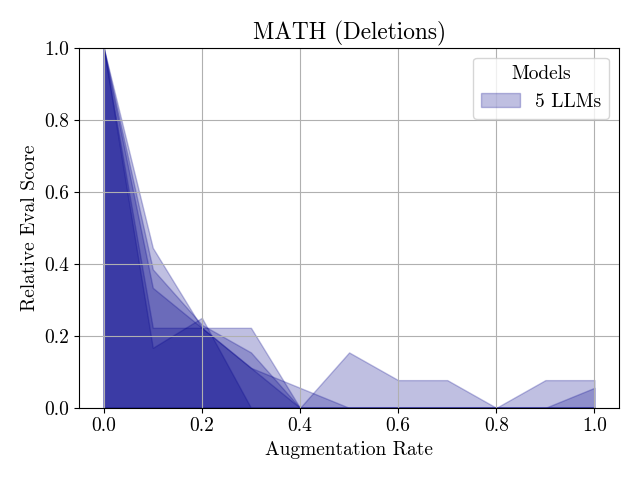}
    \\\includegraphics[width=0.23\textwidth]{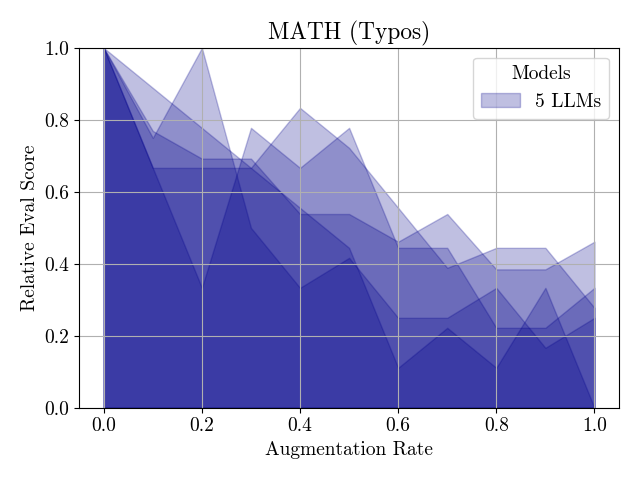}
    \caption{The average performance decay of LLMs on the MATH dataset under different obfuscation methods.}
    \label{fig:math_decay_3_methods}
\end{figure}

\section{Comparison of models and methods used}

\subsection{Initial Performance}

Despite the models relative differences in size, cost, and release date, we did not note significant differences between their performance and behaviour on either of the code-related datasets. The comparison is shown in Table~\ref{tab:baseline_models}.

\begin{table}[H]
\centering
\begin{tabular}{lccc}
\toprule
Model & OldLC & NewLC & MATH \\
\midrule
Claude 3.5      & 0.50 ± 0.12 & 0.19 ± 0.10 & 0.45 ± 0.11 \\
DeepSeek V3     & 0.65 ± 0.10 & 0.29 ± 0.11 & 0.45 ± 0.12 \\
Gemini 2.0      & 0.65 ± 0.12 & 0.40 ± 0.13 & 0.90 ± 0.08 \\
Llama 3.3       & 0.60 ± 0.11 & 0.30 ± 0.11 & 0.65 ± 0.12 \\
GPT-4o-mini     & 0.66 ± 0.11 & 0.45 ± 0.11 & 0.60 ± 0.12 \\
\bottomrule
\end{tabular}
\caption{Baseline non-obfuscated evaluation scores for each LLM across the three datasets.
}
\label{tab:baseline_models}
\end{table}

Overall, the models performed similarly to each other on the unobfuscated tasks. Gemini's performance on the MATH dataset is the only outlier found, which is not surprising as it's currently top of the benchmark leaderboard for this dataset \cite{mathleaderboard}.

\subsection{Performance Decay}

Using the previously defined 50\% decay metric of resistance to obfuscation, we report that across the 3 datasets and 5 LLMs, the obfuscation methods have seen the rates of decline presented in Table~\ref{tab:halfmax_models}

\begin{table}[H]
\centering
\begin{tabular}{lccc}
\toprule
Model & OldLC & NewLC & MATH \\
\midrule
Claude 3.5      & 0.79 ± 0.10 & 0.46 ± 0.26 & 0.14 ± 0.13 \\
DeepSeek V3     & 0.73 ± 0.13 & 0.35 ± 0.16 & 0.10 ± 0.12 \\
Gemini 2.0      & 0.60 ± 0.08 & 0.22 ± 0.10 & 0.20 ± 0.08 \\
Llama 3.3       & 0.70 ± 0.17 & 0.29 ± 0.11 & 0.21 ± 0.12 \\
GPT-4o-mini     & 0.57 ± 0.12 & 0.26 ± 0.15 & 0.24 ± 0.10 \\
\bottomrule
\end{tabular}
\caption{50\% decay point for LLMs across the datasets.}
\label{tab:halfmax_models}
\end{table}

The impact of obfuscation on the performance of the 5 LLMs was similar. While pairwise statistically significant relationships exist for the decay statistic, no single model was found to be an outlier. 

\section{Testing the error correction hypothesis}

The performance of LLMs on obfuscated tasks may be dictated by two mechanisms - legitimate error correction capabilities, and illegitimate overfitting to previosly seen tasks.

We constructed adversarial examples of tasks that aimed to explore the relative strengths of these mechanisms, testing whether the LLM will correct errors, or rather solve another question. An example of this is outlined below.

Based on the previously explored ``median of two arrays'' we create a new problem that, while using all the same key features, asks a different question that requires a different code to solve. This is presented in Figure~\ref{fig:adversarial_error_correction}.

\begin{figure}[H]
\footnotesize
\begin{center}
\begin{minipage}{0.97\linewidth}
\begin{verbatim}
Write Python code to solve the following problem:

Consider an array that consists of two concatenated 
sorted arrays of size m and n. Identify which of 
two arrays has more elements. The overall run time 
complexity should be O(log (m+n)).
Constraints:
0 <= m <= 1000
0 <= n <= 1000
1 <= m + n <= 2000
-106 <= array[i] <= 106
\end{verbatim}
\end{minipage}
\end{center}
\caption{Stylistically similar question to the ``median of two sorted arrays'', one of the first LeetCode questions \cite{leetcode_problemset}.}
\label{fig:adversarial_error_correction}
\end{figure}

In its original form, all tested LLM models are capable of solving this task correctly. We applied the obfuscation technique to test if the LLMs will still be able to solve it, or if they will eagerly pattern match to a known problem. In the obfuscated form, of the 5 queried LLMs, 3 identify this problem as ``median of two sorted arrays'', 2 produce unrelated solutions, and none identified the actual problem. We reproduced this experiment on other questions with similar outcomes.

Results like these disprove the hypothesis that the high accuracy on obfuscated tasks stems mostly from good error correction and show that the effect of gravitating towards previously-seen examples is significantly stronger.
\end{document}